\documentclass{article}

\usepackage{arxiv}

\usepackage[utf8]{inputenc} 
\usepackage[T1]{fontenc}    
\usepackage{hyperref}       
\usepackage{url}            
\usepackage{booktabs}       
\usepackage{amsfonts}       
\usepackage{nicefrac}       
\usepackage{microtype}      
\usepackage{natbib}
\usepackage{xcolor}
\usepackage{tikzsymbols}
\usepackage{fourier}
\usepackage{acronym}
\usepackage{caption}

\newcommand{\heart}{\textsuperscript{$\heartsuit$}}  
\newcommand{\club}{\textsuperscript{$\clubsuit$}} 
\newcommand{\diam}{\textsuperscript{$\diamondsuit$}} 
\newcommand{\spade}{\textsuperscript{$\spadesuit$}} 
\newcommand{\td}{\textsuperscript{\textdagger}}
\newcommand{\tdbl}{\textsuperscript{\textdaggerdbl}}

\title{On the comparability of pre-trained language models}

\author{
  Matthias~Aßenmacher \\
  Department of Statistics\\
  Ludwig-Maximilians-Universität \\
  Munich, Germany \\
  \texttt{assenmacher@stat.uni-muenchen.de} \\
  \And
  Christian~Heumann \\
  Department of Statistics\\
  Ludwig-Maximilians-Universität \\
  Munich, Germany \\
  \texttt{chris@stat.uni-muenchen.de} \\
}

\begin{document}

\maketitle

\begin{abstract}
Recent developments in unsupervised representation learning have successfully established the concept of transfer learning in NLP. Mainly three forces are driving the improvements in this area of research:\\
More elaborated architectures are making better use of contextual information. Instead of simply plugging in static pre-trained representations, these are learned based on surrounding context in end-to-end trainable models with more intelligently designed language modelling objectives. Along with this, larger corpora are used as resources for pre-training large language models in a self-supervised fashion which are afterwards fine-tuned on supervised tasks. Advances in parallel computing as well as in cloud computing, made it possible to train these models with growing capacities in the same or even in shorter time than previously established models. These three developments agglomerate in new state-of-the-art (SOTA) results being revealed in a higher and higher frequency. It is not always obvious where these improvements originate from, as it is not possible to completely disentangle the contributions of the three driving forces.\\
We set ourselves to providing a clear and concise overview on several large pre-trained language models, which achieved SOTA results in the last two years, with respect to their use of new architectures and resources. We want to clarify for the reader where the differences between the models are and we furthermore attempt to gain some insight into the single contributions of lexical/computational improvements as well as of architectural changes. We explicitly do not intend to quantify these contributions, but rather see our work as an overview in order to identify potential starting points for benchmark comparisons. Furthermore, we tentatively want to point at potential possibilities for improvement in the field of open-sourcing and reproducible research.
\end{abstract}


\section{Introduction}\label{sec:intro}

For the approaches towards most NLP tasks, researchers turn to using pre-trained word embeddings \citep{mikolov2013efficient,pennington2014glove, bojanowski2017enriching} as a key component of their models. The representations map each word of a sequence $(w_1, \dots, w_T)$ to a real valued vector of dimension $d$. A drawback of these kinds of externally learned features is that they are (i) fixed, i.e. can not be adapted to a specific domain they are used in, and (ii) context independent, i.e. there's only one embedding for a word by which it is represented in any context.\\
More recently, transfer learning approaches, as for example using convolutional neural networks (CNNs) pre-trained on ImageNet \citep{krizhevsky2012imagenet} in computer vision, have entered the discussion. Transfer learning in NLP context means pre-training a network with a self-supervised\footnote{With self-supervised learning we refer to a technique, where the labels are automatically generated from the data itself without relying on external labels.} objective on large amounts of plain text and fine-tune its weights afterwards on a task specific, labelled data set. For a comprehensive overview on the current state of transfer learning in NLP, we recommend the excellent tutorial and blog post by \cite{ruder-etal-2019-transfer}\footnote{\url{https://ruder.io/state-of-transfer-learning-in-nlp/}}.\\ 
With ULMFiT (\textbf{U}niversal \textbf{L}anguage \textbf{M}odel \textbf{Fi}ne \textbf{T}uning), \cite{howard2018universal} proposed a LSTM-based \citep{hochreiter1997long} approach for transfer learning in NLP using AWD-LSTMs \citep{merity2017regularizing}. After pre-training on a large unlabelled corpus, a task-specific layer is added to the network and the whole network is fine-tuned using labelled data. This model can be characterised as unidirectional contextual, while a bidirectionally contextual LSTM-based model was presented in ELMo (\textbf{E}mbeddings from \textbf{L}anguage \textbf{Mo}dels) by \cite{peters2018deep}. 

\begin{table}[htbp]
  \centering
    \caption{Summarizaton of the basic facts of the evaluated model architectures. Despite not being a central part of this evaluation, Word2Vec and FastText are added as baseline comparisons. With \textit{Transfer learning integration}, we try to specify to which degree the model is capable for transfer learning. We distinguish between embedding models and end-to-end trainable transfer learning models.}
    \vspace{.25cm}
  \begin{tabular}{lllll}
    \toprule
    && \multicolumn{3}{c}{Architectural Details}                   \\
    \cmidrule(r){3-5}
    Model    & Release & Architecture & Contextuality  & Transfer learning integration  \\
    \midrule
    Word2Vec & 01/2013 & FCNN           & None           & Embedding model \\
    FastText & 07/2016 & FCNN           & None           & Embedding model \\
    ULMFiT   & 01/2018 & Forward LSTM   & Unidirectional & Fully end-to-end trainable  \\
    ELMo     & 02/2018 & biLSTM         & Bidirectional  & Embedding model\\
    GPT      & 06/2018 & Transformer    & Unidirectional & Fully end-to-end trainable  \\
    BERT     & 10/2018 & Transformer    & Bidirectional  & Fully end-to-end trainable \\
    GPT2     & 02/2019 & Transformer    & Unidirectional & Fully end-to-end trainable \\
    XLNet    & 06/2019 & Autoregressive & Bidirectional  & Fully end-to-end trainable \\
             &         & Transformer    &                &                            \\
    RoBERTa  & 07/2019 & Transformer    & Bidirectional  & Fully end-to-end trainable \\
    ALBERT   & 09/2019 & Transformer    & Bidirectional  & Fully end-to-end trainable \\
    \bottomrule \\
  \end{tabular}
  \label{tab:architectures}
\end{table}

The bidirectionality in ELMo is achieved by using biLSTMs instead of AWD-LSTMs. On the other hand, ULMFiT uses a more "pure" transfer learning approach compared to ELMo, as the ELMo-embeddings are extracted from the pre-training model and are \textit{not} fine-tuned in conjunction with the weights of the task-specific architecture.\\
The OpenAI GPT (\textbf{G}enerative \textbf{P}re-\textbf{T}raining, \cite{radford2018improving}) is a model which resembles the characteristics of ULMFiT in two crucial points. It is a unidirectional language model and it allows stacking tasks specific layers on top after pre-training, i.e. it is fully end-to-end trainable. The major differences between these two models is the architecture inside the LM, where OpenAI GPT uses the Transformer architecture \citep{vaswani2017attention}.\\
Instead of processing one of the input tokens at a time, like recurrent architectures (LSTMs, GRUs) do, the Transformer takes in the whole sequence all at once. This is possible because it utilizes a variant of the \textit{Attention} mechanism \citep{bahdanau2014neural}, which allows to model dependencies without having to feed the data to the model sequentially. At the same time, the OpenAI GPT can be characterised as unidirectional model as it just takes into account the left side of the context. Its successor OpenAI GPT2 \citep{radford2019language} possesses (despite some smaller architectural changes) mainly the same model architecture and can thus also be termed as unidirectional contextual.\\
Original BERT (\textbf{B}idirectional \textbf{E}ncoder \textbf{R}epresentations from \textbf{T}ransformers, \cite{devlin2018bert}), and consequently the other two BERT-based approaches discussed here \citep{liu2019roberta,lan2019albert} as well, differ from the GPT models by the fact that they are bidirectional Transformer models. \cite{devlin2018bert} developed \textit{Masked Language Modelling} (MLM) as a special training objective which allows the use of a bidirectional Transformer without compromising the language modelling objective. XLNet \citep{yang2019xlnet} on the contrary relies on an objective which the authors call \textit{Permutation Language Modelling} (PLM) and thus also achieves to model a bidirectional context despite being an auto-regressive model. A brief overview on the characteristics of the explained models can be found in table \ref{tab:architectures}.

\section{Related work}\label{sec:related}

In their stimulating paper, \cite{raffel2019exploring} take several steps in a similar direction by trying to ensure comparability among different transformer-based models. They perform various experiments with respect to the transfer learning ability of a transformer encoder-decoder architecture by varying the pre-training objective (Different variants of denoising vs. language modelling), the pre-training resources (their newly introduced C4 corpus vs. variants thereof) and the parameter size (from 200M up to 11B). Especially, their approach of introducing a new corpus and creating subsets resembling previously used corpora like RealNews \citep{zellers2019defending} or OpenWebText \citep{gokaslan2019openweb} is a promising approach in order to ensure comparability.\\
However, their experiments do not cover an important point we trying to address in our paper:\\Focussing on only one specific architecture does not yield an answer to the question which components explain the performance differences between two models where the overall architecture differs as well (e.g. Attention-based vs. LSTM-based). \cite{yang2019xlnet} also address model comparability to some extent by performing an ablation study to compare their XLNet explicitly to BERT \citep{devlin2018bert}. In this ablation study, they train six different XLNet-based models where they modify different parts of the models in order to quantify how these design choices influence performance. At the same time they restrict themselves to an architecture of the same size as BERT-base and use the same lexical resources for pre-training. \cite{liu2019roberta} vary their RoBERTa model with respect to model size and use of pre-training resources in order to perform an ablation study aiming at comparability to BERT. \cite{lan2019albert} go even one step further with their ALBERT model by also comparing their model to BERT with regard to run time and width/depth of the model.

Despite all these experiments are highly valuable steps into the direction of better comparability, there are still no clear guidelines on which comparisons to perform in order to ensure a maximum degree of model comparability with respect to potentially influential factors.

\section{Materials and Methods}\label{methods}

First, we will present the different available corpora which were utilised for pre-training the models and compare them with respect to their size, the domain they're from and their accessibility. Subsequently, we will briefly introduce common benchmark data sets which the models are fine-tuned and evaluated on.\\
While the conceptual differences between the evaluated models have already been addressed in the introduction, the models will now be described in more detail. This is driven by the intention to emphasise differences beyond the obvious, conceptual ones.

\subsection{Training corpora}\label{sec:corpora}

We will start this chapter by briefly introducing the pre-training resources, which are commonly used. While there are some corpora that are commonly used by most of the models, some other corpora are often just used by one model in conjunction with one of the more popular ones. An overview is to be found in table \ref{tab:corpora}.

\paragraph{English Wikipedia} 
\cite{devlin2018bert} state that they used data from the English Wikipedia and provide a manual for crawling it, but no actual data set. Their data encompassed around 2.5B words. Wikipedia data sets are available in the Tensorflow \texttt{Datasets}-module\footnote{\url{https://www.tensorflow.org/datasets/catalog/wikipedia}}.

\begin{table}[htbp]
  \centering
    \caption{Pre-training resources used by the language models (sorted by release date). Concerning the \textit{Accessability}, the category \textit{Crawling Manual} can be ranked between the two other categories. In this case, the authors did not provide the data, but at least a (more or less detailed) manual for crawling the data (or similar data) oneself. The dollar signs in brackets signify the necessity of a payment in order to get access to the corpus. There's no information on RealNews \citep{zellers2019defending} and C4 \citep{raffel2019exploring} as these corpora were not used by the evaluated models.}
    \vspace{.25cm}
  \begin{tabular}{lllll}
    \toprule
    Corpus             & Source                 & Word-count\td  & Accessability        & Used by \\
    \midrule
    English Wikipedia & \cite{tensorflow}       & $\sim$ 2.500M & Fully available       & BERT; XLNet; \\
                      &                         &               &                       & RoBERTa; ALBERT \\
    CommonCrawl       & \small{\url{https://commoncrawl.org/}} & Unclear       & Fully available       & XLNet \\
    ClueWeb 2012-B    & \cite{callan2009clue09} & Unclear       & Fully available (\$\$)& XLNet \\
    Giga5             & \cite{parker2011english}& Unclear       & Fully available (\$\$)& XLNet \\
    1B Word Benchmark & \cite{chelba2013one}    & $\sim$ 830M   & Fully available       & ELMo \\
    BooksCorpus       & \cite{zhu2015aligning}  & $\sim$ 985M   & Not available         & OpenAI GPT; BERT; \\
                      &                         &               &                       & XLNet; RoBERTa; \\
                      &                         &               &                       & ALBERT \\
    Wikitext-103      & \cite{merity2016pointer}& $\sim$ 103M   & Fully available       & ULMFit  \\
    CC-News           & \cite{liu2019roberta}   & Unclear       & Crawling Manual       & RoBERTa \\
    Stories           & \cite{trinh2018simple}  & $\sim$ 7.000M\tdbl & Fully available  & RoBERTa \\ 
    WebText           & \cite{radford2019language}& Unclear     & Not available         & OpenAI GPT2 \\
    OpenWebText       & \cite{gokaslan2019openweb}& Unclear     & Fully available       & RoBERTa \\
    \bottomrule \\
  \end{tabular}
    \caption*{\textsuperscript{\textdagger} We report the word-count as given in the respective articles proposing the corpora. Note that the number of \textit{tokens} reported in  depends on the tokenization scheme used by a specific model. 
    \newline \textsuperscript{\textdaggerdbl} Stated by one of the authors on twitter:  \href{https://web.archive.org/web/20200102133529/https:/twitter.com/thtrieu\_/status/1096672446864748545}{https:/twitter.com/thtrieu\_/status/1096672446864748545}
    }
  \label{tab:corpora}
\end{table}

\paragraph{CommonCrawl} 
Among other resources, \cite{yang2019xlnet} used data from CommonCrawl. Besides stating that they filtered out short or low-quality content no further information is given. Since CommonCrawl is a dynamic database, which is updated on a monthly base, and the extracted amount of data always depends on the user, we can not provide a word count for this source in table \ref{tab:corpora}.

\paragraph{ClueWeb \citep{callan2009clue09}, Giga5 \citep{parker2011english}} 
The information about the use of ClueWeb and Giga5 is similarly sparse as for CommonCrawl (all three were used for pre-training XLNet). ClueWeb was obtained by crawling $\sim$ 2.8M web pages in 2012, Giga5 was crawled between 01/2009 and 12/2010.

\paragraph{1B Word Benchmark\protect\footnote{\url{https://ai.google/research/pubs/pub41880}} \citep{chelba2013one}} 
This corpus, actually introduced as a benchmark data set by \cite{chelba2013one} back in 2013, combines multiple data sets from the EMNLP 2011 workshop on Statistical Machine Translation\footnote{\url{http://statmt.org/wmt11/}} (WMT11). The authors normalised and tokenized the corpus and performed further pre-processing steps in dropping duplicate sentences as well as discarding words with a count below three. Additionally they randomised the ordering of the sentences in the corpus. This constitutes a corpus with a vocabulary of 793.471 words and a total word count of 829.250.940 words.

\paragraph{BooksCorpus\protect\footnote{\url{https://yknzhu.wixsite.com/mbweb}} \citep{zhu2015aligning}}
With their work from 2015, Zhu et al. introduced two corpora: the MovieBook Dataset and the BooksCorpus, with the latter one being heavily used for pre-training language models (cf. table \ref{tab:corpora}). In their work, they used the BooksCorpus in order to train a model for retrieving sentence similarity.\\
Overall, the corpus comprises 984.846.357 words\footnote{\cite{devlin2018bert} report that the BooksCorpus consists of only 800M words; we assume that the differences are attributed to potentially different pre-processing} in 74.004.228 sentences obtained from analysing 11.038 books. The vocabulary consists of 1.316.420 unique words, making the corpus lexically more diverse than the 1B Word Benchmark as it possesses a by 66\% larger vocabulary whereas having a word count which is only 19\% higher. Unfortunately it is not available for public download anymore, the authors just provide a link to the ebook-store where they scraped the corpus.

\paragraph{Wikitext-103\protect\footnote{\url{http://academictorrents.com/details/a4fee5547056c845e31ab952598f43b42333183c}} \citep{merity2016pointer}} 
\cite{merity2016pointer} emphasised the necessity for a new large scale language modelling data set by stressing the shortcomings of other corpora. They explicitly highlight the occurrence of complete articles, which allow the models to learn long range dependencies, as one of the main benefits of their corpus. This property is, according to \cite{merity2016pointer}, not given in the 1B Word Benchmark as the sentence ordering is randomised there. With a count of 103.227.021 tokens and a vocabulary size of 267.735 it is about one eighth of the 1B Word Benchmark's size concerning token count and about one third concerning the vocabulary size. Note, that there is also a smaller corpus available \footnote{Wikitext-2: \url{http://academictorrents.com/details/ac7ffa98b66427246a316a81b2ea31c9b58ea5b6}}, which is a subset of about 2\% of the size of Wikitext-103.

\paragraph{CC-News \citep{nagel2016ccnews}} 
The CC-News corpus was presented and used in \cite{liu2019roberta}. They used a web crawler proposed by \cite{hamborg2017news} to extract data from the CommonCrawl News data set \citep{nagel2016ccnews} and obtained a data set similar to the RealNews data set \citep{zellers2019defending}.

\paragraph{Stories\protect\footnote{\url{https://console.cloud.google.com/storage/browser/commonsense-reasoning/reproduce/stories_corpus}} \citep{trinh2018simple}} 
This data set is also a specific subset of the CommonCrawl data. The authors built the data based on questions in common sense reasoning tasks. They extracted nearly 1M documents, most of which are taken from longer, coherent stories (hence the name of the corpus). One of the authors stated on twitter\footnote{\url{https://twitter.com/thtrieu_/status/1096672446864748545}} that the corpus contains approximately 7B words.

\paragraph{WebText \citep{radford2019language}} 
The data set GPT2 was pre-trained on, is not publicly available and was obtained by creating "a new web scrape which emphasised document quality" \citep{radford2019language}.

\paragraph{OpenWebText\protect\footnote{\url{http://academictorrents.com/details/36c39b25657ce1639ccec0a91cf242b42e1f01db}}\citep{gokaslan2019openweb}} 
As a reaction to \cite{radford2019language} \textit{not} releasing their pre-training corpus, \cite{gokaslan2019openweb} started an initiative to emulate an open-source version of the WebText corpus.

It becomes obvious that there is a lot of heterogeneity with respect to the observed combinations of availability and the clear specification of the corpus size as word count. Some corpora specify their size in gigabytes, but do not provide a token count or a vocabulary size. Thus, we can state that there is some lack of transparency when it comes to the lexical resources used for per-training. Especially, the missing availability of the BooksCorpus is problematic as this corpus is heavily used for pre-training.

\subsection{Benchmark data sets for fine-tuning}\label{sec:benchmarks}

Besides describing pre-training resources, it is also important to have a look at the data sets which are commonly used for benchmarking fine-tuned language models and thus determine new SOTA results.

\paragraph{GLUE\protect\footnote{\url{https://gluebenchmark.com/}} \citep{wang2018glue}}

The \textit{General Language Understanding Evaluation} (GLUE) benchmark is a freely available collection of nine data sets which models can be evaluated on. It also provides a fixed train-dev-test split with held out labels for the test set, as well as a leader board which displays the top submissions and the current SOTA. The relevant metric for the SOTA is an aggregate measure of the nine single task metrics.\\
Table \ref{tab:glue} provides the basic information on the data sets included in GLUE. The benchmark includes two binary classification tasks with single-sentence inputs (CoLa [\citealp{warstadt2018neural}] and SST-2 [\citealp{socher2013recursive}]) and five binary classification tasks with inputs that consist of sentence-pairs (MRPC [\citealp{dolan2005automatically}],  QQP\footnote{\url{https://www.quora.com/q/quoradata/First-Quora-Dataset-Release-Question-Pairs}}, QNLI [\citealp{wang2018glue}], RTE [\citealp{wang2018glue}] and WNLI [\citealp{wang2018glue}]). The remaining two tasks also take sentence-pairs as input but have a multi-class classification objective with either three (MNLI [\citealp{williams2017broad}]) or five classes (STS-B [\citealp{cer2017semeval}]).

\begin{table}[htbp]
  \centering
    \caption{A brief summarizaton of the different data sets which all together form the GLUE benchmark. This table is basically a rearrangement of table 1 from \cite{wang2018glue} with slightly reduced information as it is just thought to be an overview on the different tasks and data set sizes.}
    \vspace{.25cm}
  \begin{tabular}{clllllllll}
    \toprule
& \multicolumn{2}{c}{Single-Sentence} & \multicolumn{3}{c}{Similarity/Paraphrase} & \multicolumn{4}{c}{Inference} \\
    \cmidrule(r){2-3} \cmidrule(r){4-6} \cmidrule(r){7-10}
    \textbf{task}    & CoLa  & SST-2 & MRPC & STS-B & QQP  & MNLI & QNLI & RTE  & WNLI \\
    \midrule
    \textbf{|train|} & 8.5k  & 67k   & 3.7k & 7k    & 364k & 393k & 105k & 2.5k & 634  \\
    \textbf{|test|}  & 1k    & 1.8k  & 1.7k & 1.4k  & 391k & 20k  & 5.4k & 3k   & 146  \\
    \textbf{domain}  & misc. & movies& news & misc. & social QA & misc. & wiki & news, wiki & fiction \\
    \bottomrule \\
  \end{tabular}
  \label{tab:glue}
\end{table}

\paragraph{SuperGLUE\protect\footnote{\url{https://super.gluebenchmark.com/}} \citep{wang2019superglue}}
As a reaction to human baselines being surpassed by the top ranked models, \cite{wang2019superglue} proposed a set of benchmark data sets similar to, but, according to the authors, more difficult than GLUE. On average, the size of the provided training data is smaller than in GLUE and, differently to GLUE, the data is also split in 'train', 'dev' and 'test' as in GLUE. As of the writing of this paper, there is a large difference between the use of GLUE and SuperGLUE concerning the number of models evaluated on the respective benchmark.

\begin{table}[htbp]
  \centering
    \caption{A brief summarizaton of the different data sets which all together form the SuperGLUE benchmark. This table is basically a rearrangement of table 1 from \cite{wang2019superglue} with slightly reduced information as it is just thought to be an overview on the different tasks and data set sizes.}
    \vspace{.25cm}
  \begin{tabular}{cllllllll}
    \toprule
& \multicolumn{1}{c}{Coreference} & \multicolumn{1}{c}{Disambig.} & \multicolumn{2}{c}{Inference} & \multicolumn{4}{c}{Question Answering} \\
    \cmidrule(r){2-2} \cmidrule(r){3-3} \cmidrule(r){4-5} \cmidrule(r){6-9}
    \textbf{task}    & WSC     & WiC    & RTE        & CB    & BoolQ        & COPA       & MultiRC & ReCoRD \\
    \midrule
    \textbf{|train|} & 554     & 6k     & 2.5k       & 250   & 9.4k         & 400        & 5.1k    & 101k   \\
    \textbf{|dev|}   & 104     & 638    & 278        & 57    & 3.3k         & 100        & 953     & 10k    \\
    \textbf{|test|}  & 146     & 1.4k   & 300        & 250   & 3.2k         & 500        & 1.8k    & 10k    \\
    \textbf{domain}  & fiction & misc.  & news, wiki & misc. & google, wiki & blogs, art & misc.   & news   \\
    \bottomrule \\
  \end{tabular}
  \label{tab:superglue}
\end{table}

It is considered to be more difficult than GLUE as it contains more complex tasks than just single-sentence or sentence-pair classification. SuperGLUE also features coreference resolution and question answering tasks. Unfortunately, it did not make sense to include it as a part of our model comparison, as (at the time of writing) only two of the discussed models were evaluated on SuperGLUE.

\paragraph{SQuAD\protect\footnote{\url{https://rajpurkar.github.io/SQuAD-explorer/}} \citep{rajpurkar2016squad, rajpurkar2018know}}
In its first version, the \textbf{S}tanford \textbf{Qu}estion \textbf{A}nswering \textbf{D}ataset (SQuAD) 1.1 \citep{rajpurkar2016squad} consists of 100.000+ questions explicitly designed to be answerable by reading segments of Wikipedia articles. The task is to correctly locate the segment in the text which contains the answer. A shortcoming of this task is the omission of situations where the the question is not answerable by reading the provided article. \cite{rajpurkar2018know} address this problem in SQuAD 2.0 by adding 50.000 handcrafted unanswerable questions to the SQuAD 1.1 data set. On their homepage, the authors provide a train and development set as well as an official leader board. The test set is completely held out. Instead, the participants are required to upload their models to CodaLab\footnote{\url{https://codalab.org/}}.
The SQuAD 1.1 data is, in an augmented form (termed QNLI), also part of the GLUE benchmark.

\paragraph{RACE\protect\footnote{\url{http://www.qizhexie.com/data/RACE_leaderboard.html}} \citep{lai2017race}}
The Large-scale \textbf{R}e\textbf{A}ding \textbf{C}omprehension Dataset From \textbf{E}xaminations (RACE) contains (english) exam questions for Chinese students (middle and high school). In most of the articles, where the model is evaluated on RACE, it is described to be especially challenging due to (i) the length of the passages, (ii) the inclusion of reasoning questions and (iii) the intentionally tricky design of the questions in order to test a human's ability in reading comprehension. The data set can be subdivided in RACE-M (middle school examination) and RACE-H (high school examination) and comprises a total of 97.687 questions on 27.933 passages of text. 

\subsection{Evaluated Models}\label{sec:models}

\paragraph{ULMFit \citep{howard2018universal}} 
The first "pure" transfer learning applied in NLP was ULMFiT in the beginning of 2018. The core of the model builds on the work from \cite{merity2017regularizing} as it uses AWD-LSTMs, which is a LSTM-variant that makes use of DropConnect \citep{wan2013regularization} for better regularisation and applies averaged stochastic gradient descent (ASGD) for optimization \citep{polyak1992acceleration}. This model consists of a 400 dimensional embedding layer followed by three LSTM layers, each of which encompasses 1150 hidden units. \cite{howard2018universal} stack a softmax classifier with a hidden layer size of 50 on top of this architecture for pre-training the model. This final layer is complemented by a task specific final layer during fine tuning. The vocabulary size is limited to 30k words as in \cite{johnson2017deep}.\\
In contrast to the other models discussed in this paper, ULMFiT was not evaluated on the GLUE benchmark but on several other data sets (IMDb [\citealp{maas2011learning}], TREC-6 [\citealp{voorhees1999trec}], Yelp-bi, Yelp-full, AG's news, DBpedia [all \citealp{zhang2015character}]).

\begin{table}[htbp]
  \centering
    \caption{An overview on the data sets which ULMFit was fine-tuned and evaluated on. It is an extension of table 1 \citep{howard2018universal}, adding information on the size of the test set and the domain. All six tasks are classification tasks, where the target variables have between 2 and 14 classes.}
    \vspace{.25cm}
  \begin{tabular}{cllllllll}
    \toprule
& \multicolumn{1}{c}{Question} & \multicolumn{3}{c}{Sentiment} & \multicolumn{2}{c}{Topic}  \\
    \cmidrule(r){2-2} \cmidrule(r){3-5} \cmidrule(r){6-7}
    \textbf{task}    & TREC-6  & IMDb    & Yelp-bi   & Yelp-full & AG's news & DBpedia  \\
    \midrule
    \textbf{|train|} & 5  k    & 25k     & 560k      & 650k      & 120k      & 560k     \\
    \textbf{|test|}  & 0.5k    & 25k     & 38k       & 50k       & 7.6k      & 70k      \\
    \textbf{domain}  & open-domain & movies  & social QA & social QA & news      & wiki     \\
    \bottomrule \\
  \end{tabular}
  \label{tab:ulmfit}
\end{table}

\paragraph{ELMo \citep{peters2018deep}}
As already stated in section \ref{sec:intro}, ELMo differs from ULMFit with respect to its usability for transfer learning. The pre-trained ELMo-embeddings are plugged in at the lowest layer of an arbitrary NLP model in order to use them for a downstream task\footnote{The authors also mention that additionally adding the ELMO-embeddings at one of the final layers might improve performance for some architectures and tasks}. In case of ELMo this means the following: As ELMo consists of multiple biLSTM layers, one can extract multiple intermediate-layer representations from the model. These representations are used for computing a (task-specific) weighted combination, which is concatenated with static context-independent word embeddings. So the model weights of ELMo are not updated during the training of the downstream model, but only the weights, which are learned for combining the intermediate-layer representations from ELMo, are. \cite{peters2018deep} evaluate an ELMo-based model on SQuAD and other tasks, but when it comes to GLUE there are multiple ELMo-based architectures available on the leaderboard\footnote{\url{https://gluebenchmark.com/leaderboard}}. Thus, here we will concentrate on the best-performing ELMo-based model, a BiLSTM-model with Attention \citep{wang2018glue}. 

\paragraph{OpenAI GPT \citep{radford2018improving}}
The OpenAI GPT is a pure attention-based architecture the does not make use of any recurrent layers. Pre-training is performed by combining Byte-Pair encoded \citep{sennrich2015neural} token embeddings with learned position embeddings, feeding them into a multi-layer transformer decoder architecture with a standard language modelling objective. By using a decoder architecture the model does at each step only have access to the preceding tokens in the sequence. Thus, the GPT model is a unidirectional attention-based architecture. Fine-tuning was, amongst others, performed on the nine tasks that together form the GLUE benchmark.

\paragraph{BERT \citep{devlin2018bert}}
This model can be seen as a reference point for everything that came thereafter. Similar to GPT it uses Byte-Pair Encoding (BPE) with a vocabulary size of 30k. By introducing the MLM training objective, the authors were able to combine deep bidirectionality with the self-attention mechanism for the first time. In addition to the MLM objective it also utilizes as next-sentence prediction (NSP) objective, the usefulness of which has been debated in other research papers \citep{liu2019roberta}. The BERT-BASE model consists of 12 bidirectional transformer-encoder blocks (24 for BERT-LARGE) as described in \cite{vaswani2017attention} with 12 (16 respectively) attention heads per block and an embedding size of 768 (1024 respectively). The need to better understand the behaviour of these huge networks even constituted a new field of research called \textit{BERTology}, aiming at explaining the inner workings of BERT-based models.

\paragraph{OpenAI GPT2 \citep{radford2019language}}
With GPT2, the OpenAI team published a scaled-up version of GPT in 2019. Compared to its predecessor, it contains some smaller changes concerning the placement of layer normalisation and residual connections. Overall, there are four different versions of GPT2 with the smallest one being equal to GPT, the medium one being of similar size as BERT-LARGE and the xlarge one being released as the actual GPT2 model with 1.5B parameters.

\paragraph{XLNet \citep{yang2019xlnet}}
In order to overcome (what they call) the \textit{pretraining-finetune discrepancy}, which is a consequence of BERT's masking approach, and to simultaneously include bidirectional contexts, \cite{yang2019xlnet} propose the \textit{PLM} objective for their XLNet. They use two-stream self-attention for preserving the position information of the token to be predicted, which would otherwise be lost due to the permutation of the sequence. While the first of the two streams (\textit{content stream attention}) resembles the standard self-attention from a transformer-decoder, the other stream (\textit{query stream attention}) doesn't allow the token to see itself but just the preceding tokens of the permuted sequence.

\paragraph{RoBERTa \citep{liu2019roberta}}
With RoBERTa (short for \textbf{R}obustly \textbf{o}ptimized \textbf{BERT} \textbf{a}pproach), \cite{liu2019roberta} introduce an exact (architectural) replicate of BERT with tuned hyperparameters and a larger corpus used for pre-training. The masking strategy for pre-training is changed from static (masking once during pre-processing) to dynamic (masking every sequence just before feeding it to the model), the additional NSP objective is removed, the BPE-level vocabulary is adjusted and increased to 50k and RoBERTa is trained on larger batches than BERT. All of these adjustments improve performance of the model and make it competitive to the previously SOTA results of XLNet.

\paragraph{ALBERT \citep{lan2019albert}}
By addressing the steady increase of the model size as a potential problem, ALBERT (short for \textbf{A} \textbf{L}ite \textbf{BERT}) goes into another direction compared to most of post-BERT architectures. \cite{lan2019albert} apply parameter-reduction techniques in order to train faster models with lower memory demands that, at the same time, yield a comparable performance to SOTA models. In our work we will always refer to ALBERT-XXLARGE, which is the best performing ALBERT model. Note, that also the much smaller ALBERT models yielded results comparable to or even better than BERT. 

\section{Model comparison}\label{sec:comparison}

The two tables below will try to give a comprehensive overview on the differences of the previously discussed model architectures. While table \ref{tab:arch-details} will only attempt to give an overview on the amount of computation that was needed to train a given architecture on a given corpus, we will directly try to relate model architecture and size as well as usage of lexical resources to model performance in table \ref{tab:arch-performance}.

\begin{table}[htbp]
  \centering
    \caption{Summarizaton of the basic facts of the evaluated transfer learning model architectures. Word2Vec, FastText and ELMo are not included as these are no end-to-end trainable models, meaning that the model size also depends of the used model after obtaining the embeddings. The parameter size of ULMFiT is assumed to be the larger value from \cite{merity2017regularizing}, since \cite{howard2018universal} use plain AWD-LSTMs with a vocabulary size of 30k tokens like \cite{johnson2016convolutional,johnson2017deep}. Values for GPT2-XLARGE are taken from \cite{strubell2019energy}.}
    \vspace{.25cm}
  \begin{tabular}{llllll}
    \toprule
                 & \multicolumn{3}{c}{Compute}                   & \multicolumn{2}{c}{Resources}  \\
    \cmidrule(r){2-4}\cmidrule(r){5-6}
    Model        & Computational Resources     & Training time   & pfs-days \td & size       & lexical\\
    \midrule
    ULMFiT       & \texttt{NA}                 & \texttt{NA}     & \texttt{NA}  &    33M     & 0.18GB \\
    GPT          & 8 GPUs (P600)               & $\sim$ 30 days  & 0.96         &   117M     & < 13GB \\
    BERT-BASE    & 4 Cloud TPUs (16 chips)     & $\sim$ 4 days   & 0.96 [2.24] \tdbl  &   110M     & 13GB \\
    BERT-LARGE   & 16 Cloud TPUs (64 chips)    & $\sim$ 4 days   & 3.84 [8.96] \tdbl  &  340M      & 13GB \\
    GPT2-MEDIUM  & \texttt{NA}                 & \texttt{NA}     & \texttt{NA}  &   345M     & 40GB \\
    GPT2-XLARGE  & 8 v3 Cloud TPUs (32 chips)  & $\sim$ 7 days   & 7.84         & 1.500M     & 40GB \\
    XLNet-LARGE  & 128 v3 Cloud TPUs (512 chips)& $\sim$ 2.5 days& 44.8         &   340M     & 126GB \\
    RoBERTa      & DGX-1 GPUs (8x32GB V100)    & \texttt{NA}     & \texttt{NA}  &   360M     & 160GB \\
    ALBERT       & 64 -- 1024 v3 Cloud TPUs    & \texttt{NA}     & \texttt{NA}  &   233M     & 16GB \\
    \bottomrule \\
  \end{tabular}
    \caption*{\textsuperscript{\textdagger} Estimation according to the formula proposed on \url{https://openai.com/blog/ai-and-compute/}: 
    \newline $\texttt{pfs-days}\; =\; \texttt{number of units}\; \times \; \texttt{PFLOPS}/\texttt{unit}\; \times \; \texttt{days trained}\; \times \; \texttt{utilization}$, with an assumed \texttt{utilization} of one third. Information on \texttt{PFLOPS}/\texttt{unit} for TPUs from \url{https://cloud.google.com/tpu/}.
    \newline \textsuperscript{\textdaggerdbl} We provide two numbers here, as \cite{devlin2018bert} do not specify whether they use \texttt{v2} or \texttt{v3} TPUs. The first number assumes the use of \texttt{v2} TPUs, the one in square brackets assumes use of \texttt{v3} TPUs.
    }
  \label{tab:arch-details}
\end{table}

One thing that we can learn from table \ref{tab:arch-details} is the unfortunate lack of details when it comes to reporting the computational resources used for training the models. While \cite{howard2018universal} do not provide any information at all on the computational resources utilised for pre-training ULMFiT, the other articles are also not over-informative when it comes to reporting them. Unfortunately, there are no clear guidelines on how to appraise resource consumption when it comes to evaluating and comparing models. This may be partly attributed to the rapidly growing hardware possibilities due to modern cloud computing architectures, but in our opinion it should nevertheless be accounted for, since it may pose environmental issues \citep{strubell2019energy} and also limits portability to smaller devices.

The second thing is that it is also important to consider the differences displayed in the tables \ref{tab:arch-details} and \ref{tab:arch-performance} when comparing the model performances. When comparing two models of approximately the same size (e.g. BERT-BASE versus GPT), it seems to be obvious that the superior performance of BERT-BASE originates purely from its more elaborated model architecture (cf. table \ref{tab:architectures}) because of the similar size. But one should also be aware of the larger pre-training resources (BERT-BASE uses at least twice as much data for pre-training) as well as the unknown differences in usage of computing power. We estimated the amount of compute used by a model as the \textit{pfs-days}, resulting in an estimation for BERT-BASE being not less than the one for GPT.\\ 
Another aspect which should not be ignored when evaluating performance is the use of ensemble models. As can be seen in the first column of table \ref{tab:arch-performance}, the three ensemble models seem to outperform both of the BERT models by a large margin. Only parts of these differences may be attributed to the model architecture, as the ensembling as well as the larger pre-training resources might also give an advantage to these models. As there are unfortunately no \textit{single} model performance values available for XLNet, RoBERTa and ALBERT on the official GLUE leaderboard, we also compare the single model performances from \cite{lan2019albert} obtained on the dev sets (WNLI excluded). From this comparison we can get a good impression of how high the contribution of model ensembling might be: The difference between BERT-LARGE and the XLNet ensemble in the official scores (7.9 percentage points) is more than twice as high as the difference on the dev score (3.4 percentage points).\\
In order to address the differences in size of the pre-training resources, \cite{yang2019xlnet} make the extremely insightful effort to compare a BASE variant of XLNet to BERT-BASE (same size and same pre-training resources). While the F1 score on the SQuAD v2.0 dev set is still remarkably higher than for BERT-BASE (almost comparable to BERT-LARGE) it does not show a large improvement on the RACE test set anymore (which might have been expected due to the large improvement of XLNet-LARGE over both BERT models). 

\begin{table}[htbp]
  \centering
    \caption{Performance of different models on GLUE, SQuAD and RACE as well as model size and resource usage \textit{compared to BERT-BASE} (except for GLUE dev set performance, where \textit{BERT-LARGE} is the reference). Performance differences on the benchmark data sets are given in percentage points, while the differences in size/resources are given as factors, e.g. BERT-LARGE has 3.1 times the size of BERT-BASE and performs 2.2 percentage points better on GLUE. We omit SuperGLUE in this table as of the time of writing only BERT and RoBERTa were evaluated on it. ULMFiT and OpenAI GPT2 are also omitted as there are no performance values on these data sets publicly available. Highest improvements over the reference model in bold. For ELMo we do not provide a model size, since the performance values are from two different models (cf. section \ref{sec:models}).
    \newline Displayed performance measures are Matthews Correlation (GLUE), F1 score (SQuAD) and Accuracy (RACE).}
    \vspace{.25cm}
  \begin{tabular}{llllllll}
    \toprule
    & \multicolumn{2}{c}{GLUE} & \multicolumn{2}{c}{SQuAD} & \multicolumn{1}{c}{RACE}& \multicolumn{2}{c}{Resources} \\
    \cmidrule(r){2-3}\cmidrule(r){4-5}\cmidrule(r){6-6}\cmidrule(r){7-8}
    Model      & leaderboard    & dev\heart   & v1.1 (dev) & v2.0 (dev)    & test             & size        & lexical   \\
    \midrule\midrule
    BERT-BASE  &{\it 78.3}      & --          & {\it 88.5}    & {\it 76.3} \club & {\it 65.0} \diam & {\it 110M}  & {\it 13GB} \\
    \midrule 
    ELMo-based & - 8.3          & --          & - 2.9         & --               & --               &  --           & -- \\
    GPT        & - 5.5          & --          & --            & --               & - 6.0            &  1.1x         & < 0.5x \\
    BERT-LARGE & + 2.2          & {\it 84.05} & + 2.4         & + 5.6            & + 7.0 \diam      &  3.1x         & 1.0x \\
    XLNet-BASE & --             & --          & --            & + 5.03           & + 1.05           & $\sim$ 1.0x   & 1.0x \\
    XLNet-LARGE& + 10.1 \spade  & + 3.39      & + 6.0         & + 12.5           & + 16.75          &  3.1x         & 9.7x \\
    RoBERTa    & + 10.2 \spade  & + 5.19      & {\bf + 6.1 }  & + 13.1           & + 18.2           &  3.3x         & 12.3x \\
    RoBERTa-BASE   & --         & + 2.30      & --            & --               & --               &  1.0x         & 12.3x \\
    RoBERTa \tdbl  & --         & + 3.79      & + 5.1         & + 11.0           & --               &  3.3x         & 1.2x \td\\
    ALBERT     & {\bf + 11.1} \spade  & {\bf + 5.91} & + 5.6  & {\bf + 13.9}     & {\bf + 21.5}     &  2.1x         & 1.2x \td\\
    \bottomrule \\
  \end{tabular}
    \caption*{\textsuperscript{$\spadesuit$} Ensemble performance; No single model performance available
    \newline \textsuperscript{$\heartsuit$} Own calculations based on \cite{lan2019albert} table 13; WNLI is omitted
    \newline \textsuperscript{$\clubsuit$} Result for BERT-BASE on SQuAD v2.0 is taken from \cite{yang2019xlnet} table 6
    \newline \textsuperscript{$\diamondsuit$} Result for BERT-BASE on RACE is taken from \cite{zhang2019dual} table 2
    \newline \textsuperscript{\textdagger} \cite{liu2019roberta} and \cite{lan2019albert} specify the BooksCorpus + English Wikipedia as 16GB  
    \newline \textsuperscript{\textdaggerdbl} This variant of RoBERTa uses only BooksCorpus + English Wikipedia for pre-training
    }
  \label{tab:arch-performance}
\end{table}

The comparability of RoBERTa from the GLUE leaderboard (model ensemble and larger pre-training resources) to BERT-LARGE is again limited, but the authors performed several experiments in order to show the usefulness of their model optimisations. When pre-training BERT-LARGE and a single RoBERTa model on comparable lexical resources (BooksCorpus + English Wikipedia; 13GB for BERT vs. 16GB for RoBERTa), the RoBERTa model still shows a significant improvement over BERT-LARGE, even if it decreases somewhat in size (compared to the difference between BERT-LARGE and the ensemble model). In another ablation study, \cite{liu2019roberta} train a BASE variant of RoBERTa on their larger pre-training resources. Even though comprising only about one third of the size of BERT-LARGE, the larger pre-training corpus in conjunction with the optimised training leads to a slightly better performance on the GLUE dev set (without WNLI). Unfortunately we cannot compare RoBERTa-BASE to BERT-BASE, as we neither have the "official" leaderboard score for RoBERTa-BASE nor the "in-official" dev set score for BERT-BASE.

\begin{table}[htbp]
  \centering
    \caption{Performance of BERT-LARGE and XLNet-LARGE on the benchmark data sets used by \cite{howard2018universal} as well as model size and resource usage \textit{compared to ULMFiT}. Specification of the differences are displayed as in table \ref{tab:arch-performance}, highest improvements over the reference model in bold. Note that we report accuracies here, as opposed to \cite{howard2018universal} and \cite{yang2019xlnet}, in order to provide a more similar interpretation of these values compared to the values in table \ref{tab:arch-performance}. Displayed performance measures are Accuracy for all tasks.}
    \vspace{.25cm}
  \begin{tabular}{llllllll}
    \toprule
    & \multicolumn{3}{c}{Sentiment} & \multicolumn{2}{c}{Topic} & \multicolumn{2}{c}{Resources} \\
    \cmidrule(r){2-4} \cmidrule(r){5-6} \cmidrule(r){7-8}
    Model       & IMDb        & Yelp-bi     & Yelp-full   & AG's news   & DBpedia     & size      & lexical \\
    \midrule\midrule
    ULMFiT      & {\it 95.40}  & {\it 97.84}  & {\it 70.02}  & {\it 94.99}  & {\it 99.20}  & {\it 33M} & {\it 0.18GB} \\
    \midrule 
    BERT-LARGE  & + 0.09       & + 0.27       & + 0.66       & --           & + 0.16       & 10.3x     & 72.2x \\
    XLNet-LARGE & {\bf + 0.81} & {\bf + 0.61} & {\bf + 2.28} & {\bf + 0.52} & {\bf + 0.18} & 10.3x     & 222.2x \\
    \bottomrule \\
  \end{tabular}
  \label{tab:ulmfit-performance}
\end{table}

In order to also set the results of ULMFiT into context, we present the results published by \cite{yang2019xlnet} alongside with the information on model size and use of lexical resources in table \ref{tab:ulmfit-performance}. Despite being much larger and utilising some orders of magnitude larger corpora for pre-training, both BERT-LARGE and XLNet-LARGE do not exhibit that large improvements over the performance of ULMFiT. This might partly originate from the simplicity (compared to GLUE \& co.) of the tasks, but partly also from the already achieved high performances where no extremely large improvements are possible anymore.

\section{Discussion}\label{sec:discussion}

This chapter reflects the main takeaways from the above comparisons and tries to raise some issues for future research practices. We do not claim to have a solution to these potentially problematic aspects but think that these points are highly debatable.

\paragraph{Why no benchmark corpus for pre-training?}
It is good and well-established practice to use benchmark data sets like GLUE, SuperGLUE (not yet used that often), SQuAD and RACE for comparing the performance of pre-trained language models on different types of NLP/NLU tasks. Many recently published articles \citep{liu2019roberta,yang2019xlnet,lan2019albert} perform (partly extensive) ablation studies controlling for pre-training resources in order to make (versions of) their models comparable to BERT as "benchmark model", which is really important as it helps the reader to get an intuition for the impact of pre-training resources. Nevertheless, it is unfortunately not perfect due to two critical issues: (i) BERT (and all the other models consequently as well) make use of the BooksCorpus \citep{zhu2015aligning} which is not publicly available and (ii) this only leads to model comparisons in a low pre-training resource environment (compared to more recent models) and yields no insight on the behaviour of the reference model (e.g. BERT) in a high(er) pre-training resource context. So we view statements of the type \textit{"Model architecture A is superior to model architecture B on performing task X."} somewhat critical and would propose to phrase it in a way comparable to the following statement: \textit{"Model architecture A is superior to model architecture B on performing task X, when pre-trained on a small/large corpus of low/high quality data from domain Y for time Z."}  

\paragraph{Why no standardised description of (computational) resources?}
When writing this article, it sometimes turned out difficult to really get one (measure) for how much compute was used to pre-train the model described in an article. In our opinion, this is not a carelessness of the authors but rather the lack of a clear reporting standard. We found ourselves confronted with the following situations:
\begin{enumerate}
    \item[a)] No information at all \citep{radford2019language}
    \item[b)] Information on the used hardware \citep{liu2019roberta,lan2019albert}
    \item[c)] Information on the used hardware and training time \citep{devlin2018bert,yang2019xlnet}
    \item[d)] Calculation of a standardised measure \citep{radford2018improving}\footnote{The calculation was not published as part of the article but is to be found in a corresponding blog post:\\\url{https://openai.com/blog/language-unsupervised/}}
\end{enumerate}
While situation \textit{a)} is clearly unsatisfactory and should be avoided, scenarios \textit{b)} and \textit{c)} basically provide (almost) all of the necessary information but miss out on going the last final step to scenario \textit{d)} where the reporting would reach universal comparability across different articles. A quite nice and intuitive way was also proposed on the OpenAI-blog\footnote{\url{https://openai.com/blog/ai-and-compute/}} for estimating the GPU time needed for model training. This is of course not as exact as a computation based on the counts of operations in a model, but requires on the other hand no deep insight into the model architecture and is thus applicable to a a wide range of architectures without much effort.

\paragraph{Shouldn't performance be evaluated in relation to size and resource consumption?}
As larger models have a higher capacity for learning good representations and using larger pre-training resources should also improve their quality, varying these two components simultaneously with the model architecture might lead to interference between the individual influences on model performance. So the intent of this aspect has a slight overlap with the question posed above, but while the above is more or less about introducing some kind of reference, this is about carefully varying and evaluating the effects of different parts of the model.

\section{Conclusion}\label{sec:conclusion}

As can be seen from the above analysis, there is a clear lack of a concise guideline for \textit{fair} comparisons of large pre-trained language models. It is not sufficient to just rank models by their performance on the common benchmark data sets as this does not take into account all the other factors mentioned in this analysis.

\begin{table}[htbp]
  \centering
    \caption{Proposal of starting points when thinking about reporting standards for pre-trained LMs. We categorise the reporting of the \textit{experimental time} and the \textit{benchmark performance of the un-tuned model} as not easily feasible, as one has to be aware of these standards in order to track the time of all experiments. Also, defining what is an "un-tuned" version is not always that simple. With "un-tuned" we mean \textit{not further tuned during pre-training}.}
    \vspace{.25cm}
  \begin{tabular}{llllc}
    \toprule
    & \multicolumn{3}{c}{Evaluation}                   \\
    \cmidrule(r){2-4}
    Reporting Standard & Feasibility & Current      & Relevance for   & OK? \\
                       &             & realisation  & reproducability & \\
    \midrule
    \textbf{Model architecture}      & Easy      & Every article      & Crucial   & \checkmark \\
    \textbf{Number of parameters}    & Easy      & Most articles      & Crucial   & \checkmark \\
    \textbf{Hyperparameters}         &           & & &  \\
     -- Tuning method                & Easy      & Some articles      & High      & \danger  \\
     -- Tuning time                  & Easy      & No article         & Medium    & \danger \\
    \textbf{Experimental time}       & Difficult & No article         & Low       & \danger  \\
    \textbf{Computational resources} & Easy      & Most articles      & High      & \danger  \\
    \textbf{Training time}           & Easy      & Most articles      & Medium    & \danger  \\
    \textbf{Lexical resources}       &           & & &   \\
     -- Information                  & Easy      & Every article      & Crucial   & \checkmark  \\
     -- Availability                 & Difficult & Some articles      & Crucial   & \danger  \\
    \textbf{Benchmark performance}   &           & & &   \\
     -- Un-Tuned \textbf{single} model & Difficult & No article       & Low       & \danger  \\
     -- Tuned \textbf{single} model  & Easy      & Every article      & Crucial   & \danger  \\
    \bottomrule \\
  \end{tabular}
  \label{tab:standards}
\end{table}

A further aspect (which is not explicitly addressed here) is the reporting of resources (time and compute) spent on model development, including all experimental runs and trials, and hyperparameter tuning during pre-training. In our opinion, this is important with respect to two facets: On the one hand side it is important to take into account energy and environmental considerations when training deep learning models \citep{strubell2019energy}, on the other hand it is also a signal to the reader/user for how difficult it is to train (and to fine-tune) the model. This might have implications for the usage of a model as transfer learning model for diverse downstream tasks. Models that have already been tuned to a high degree during pre-training to reach a certain level of performance, have, in the long run, maybe less potential for further improvements than models which do so without much hyperparameter tuning.\\
Taking all these considerations into account, we want to tentatively propose starting points (cf. table \ref{tab:standards}) for defining reporting standards which are globally accepted and applied when it comes to comparing pre-trained language models. We carefully try to categorise the different facets according to feasibility (\textit{How much effort does it take to report this?}), current realisation (\textit{How many research papers are reporting this?}) and their relevance for reproducible research (\textit{How crucial is this for performing reproducible research?}). All these categorisations are of more or less subjective nature due to the fact that they cannot be quantified and are based on just a handful of the most influential research papers.\\
We are aware of the fact, that it might take a large collective effort in order to establish some set of standards but we think that it is an absolutely crucial step to describe all the aspects we mentioned in a way that is as transparent as possible in order to foster replicability and reproducability.

\bibliography{references} 
\bibliographystyle{apalike}  

\vfill

\section*{List of abbreviations}

\begin{acronym} 
\acro{awd}[AWD]{Averaged stochastic gradient decent weight-dropped}
\acro{bilstm}[biLSTM]{bi-directional Long short-term memory} 
\acro{bpe}[BPE]{Byte-Pair Encoding}
\acro{cnn}[CNN]{Convolutional neural network}
\acro{fcnn}[FCNN]{Fully connected neural network} 
\acro{gru}[GRU]{Gated recurrent unit}
\acro{lstm}[LSTM]{Long short-term memory} 
\acro{mlm}[MLM]{Masked Language Modelling} 
\acro{nlp}[NLP]{Natural Language Processing}
\acro{nlu}[NLU]{Natural Language Understanding} 
\acro{plm}[PLM]{Permutation Language Modelling} 
\acro{sota}[SOTA]{State-of-the-art}
\end{acronym}
\end{document}